\documentclass{article}

\usepackage{PRIMEarxiv}

\usepackage[utf8]{inputenc} 
\usepackage[T1]{fontenc}    
\usepackage{hyperref}       
\usepackage{url}            
\usepackage{booktabs}       
\usepackage{amsfonts}       
\usepackage{nicefrac}       
\usepackage{microtype}      
\usepackage{lipsum}
\usepackage{fancyhdr}       
\usepackage{graphicx}       
\graphicspath{{media/}}     

\usepackage{amsmath}
\usepackage{amssymb}
\usepackage{mathtools}
\usepackage{amsthm}
\usepackage{makecell}
\usepackage{algorithm}
\usepackage{algorithmic}
\usepackage{color}

\usepackage[capitalize,noabbrev]{cleveref}
\title{Coefficient Decomposition for Spectral Graph Convolution
}

\author{
  Feng Huang, Wen Zhang \\
}

\begin{document}
\maketitle

\begin{abstract}
Spectral graph convolutional network (SGCN) is a kind of graph neural networks (GNN) based on graph signal filters, and has shown compelling expressivity for modeling graph-structured data. Most SGCNs adopt polynomial filters and learn the coefficients from the training data. Many of them focus on which polynomial basis leads to optimal expressive power and models' architecture is little discussed. In this paper, we propose a general form in terms of spectral graph convolution, where the coefficients of polynomial basis are stored in a third-order tensor. Then, we show that the convolution block in existing SGCNs can be derived by performing a certain coefficient decomposition operation on the coefficient tensor. Based on the generalized view, we develop novel spectral graph convolutions CoDeSGC-CP and -Tucker by tensor decomposition CP and Tucker on the coefficient tensor. Extensive experimental results demonstrate that the proposed convolutions achieve favorable performance improvements.  
\end{abstract}


\section{Introduction}
\label{Introduction}

In recent years, Graph neural networks (GNNs) have become prevailing in graph learning due to their remarkable performance on a variety of tasks such as node classification \cite{GCN}, graph classification \& regression \cite{morris2019weisfeiler} and link prediction \cite{zhang2018link}. As graphs that represent relational and structured data are ubiquitous in real world, GNNs have been widely applied in numerous fields, including social analysis \cite{qiu2018deepinf, li2019encoding}, drug discovery \cite{rathi2019practical, jiang2021could}, traffic forecasting \cite{chen2019gated,JIANG2022117921}, network medicine \cite{xiong2021multimodal,fu2022mvgcn} and recommendation system \cite{wu2022graph,huang2021mixgcf}. Spectral GNNs, also known as spectral graph convolutional networks (SGCNs), are a class of GNNs that generalize convolutional neural networks (CNNs) to graph domain \cite{ChebNet,bo2023survey}. Their convolution operators are defined on the basis of graph Fourier transform and parameterized as graph signal filters (a function with regard to graph Laplacian spectrum) in the spectral domain. Accordingly, SGCNs have solid theoretical foundation and interpretability based on spectral graph theory. 

Most SGCNs are often designed to approximate the graph signal filters by learnable polynomials of graph Laplacian spectrum to avoid the large computational burden on the eigendecomposition. Typically, they fix a predefined polynomial basis and learn the coefficients from the training data, by which, theoretically, arbitrary filters can be obtained. ChebNet \cite{ChebNet} parameterizes the graph filter by Chebyshev polynomial. GCN \cite{GCN} only uses the first two terms of the Chebyshev polynomial and achieves a new milestone. SGC \cite{SGC} simplifies a multi-layer GCN by removing nonlinear activation functions. APPNP \cite{APPNP}, S$^2$GC \cite{SSGC} and GPR-GNN \cite{GPR_GNN} straightforwardly use Monomial basis, while BernNet \cite{BernNet} and JacobiConv \cite{JacobiConv} respectively use the non-negative Bernstein basis and orthogonal Jacobi basis. ChebNetII \cite{ChebNetII} utilizes Chebyshev interpolation to enhance the original Chebyshev polynomial approximation. 


This work focuses on unifying these spectral graph convolutions built upon polynomial filters. We first revisit model architectures of many spectral GNNs. They can be categorized into three types: multi-layer, linear and hybrid GNNs, which, respectively, refer to stacking multiple convolution layers, like ChebNet and GCN; only containing linear transformation and propagation without any nonlinear parts, like JacobiConv; and containing nonlinear part and spectral convolution part, like GPR-GNN, BernNet and ChebNetII. Then, each layer of multi-layer GNNs, the whole architecture of linear GNNs and the spectral convolution part of the hybrid GNNs can be unified as different variants of one spectral graph convolution layer (SGCL) with $K$-order polynomials and full coefficients (each channel of input graph signals has multiple learnable filters). The architectures of these variants can be deduced by certain coefficient decomposition operations on the coefficient tensor that stores all coefficients. 

With the above understandings, we further propose two kinds of SGCL, named CoDeSGC-CP and -Tucker, which are respectively derived by performing CP and Tucker decomposition on the coefficient tensor. In computational experiments on node classification task, our proposed models perform better than state-of-the-art method JacobiConv on 8 out of 10 real-world datasets.

\section{Preliminaries}
\label{preliminaries}

We denote matrices by boldface capital letters, e.g., $\boldsymbol{X}$. Vectors are denoted by boldface lowercase letters, e.g., $\boldsymbol{x}$, which is generally viewed as column vectors while a row vector is denoted as $\boldsymbol{x}^\top$. The $i$th column of $\boldsymbol{X}$ is typically denoted as $\boldsymbol{x}_i$ and the $j$th column of $\boldsymbol{X}^\top$ is denoted as $\boldsymbol{x}_{j:}$; the $i$th element of vector $\boldsymbol{x}$ is $x_i$ and the element $(i,j)$ of matrix $\boldsymbol{X}$ is $x_{ij}$. Tensors are multidimensional arrays and the order of a tensor is the number of dimensions, also known as ways or modes. Note that vectors are first-order tensors, matrices are second-order tensors, and for distinction, higher-order tensors are denoted by boldface script letters, e.g., $\boldsymbol{\mathcal{X}}$. Specifically, in a third-order tensor $\boldsymbol{\mathcal{X}}\in \mathbb{R}^{I\times J \times K}$, the element $(i,j,k)$ is denoted by $x_{ijk}$ or $x_{ij}^k$, the mode-1 fibers defined by fixing indices apart from the first one are denoted by $\boldsymbol{x}_{:jk}\in \mathbb{R}^I$ and the frontal slices defined by fixing the third index are denoted by $\boldsymbol{X}_k \in \mathbb{R}^{I \times J}$; one can unfold a tensor into a matrix by flattening the mode-1 fibers under a certain ordering and the matrix is denoted as $\boldsymbol{X}_{(1)}\in\mathbb{R}^{I\times JK}$. We can use $\operatorname{reshape}(\boldsymbol{X}_{(1)}, (I,J,K))$ to reverse the unfolding operation and return $\boldsymbol{\mathcal{X}}$. We use $\operatorname{diag()}$ to convert a vector into a diagonal matrix. There exist many types of multiplication between tensors (including matrices and vectors). $\circ$ represents the vector outer product. Specifically, the outer product of three vectors results in an third-order tensor, i.e., $\boldsymbol{\mathcal{X}}=\boldsymbol{a}\circ \boldsymbol{b} \circ \boldsymbol{c}$, where $\boldsymbol{a}\in\mathbb{R}^{I}$, $\boldsymbol{b}\in\mathbb{R}^{J}$ and $\boldsymbol{c}\in\mathbb{R}^{K}$; $\boldsymbol{\mathcal{X}}\in \mathbb{R}^{I\times J \times K}$ and each element is the product of the corresponding vector elements: $x_{ijk}=a_{i}b_{j}c_{k}$. The $n$-mode product of an $N$th-order tensor $\boldsymbol{\mathcal{X}}\in \mathbb{R}^{I_1\times I_2 \times \cdots \times I_N}$ with a matrix $\boldsymbol{M}\in \mathbb{R}^{J \times I_n}$ is denoted by $\boldsymbol{\mathcal{X}}\times_n \boldsymbol{M}$, resulting in another $N$th-order tensor $\boldsymbol{\mathcal{Y}}$ of size $I_1\times \cdots \times I_{n-1} \times J \times I_{n+1} \times \cdots \times I_N$. In element-wise form, $n$-mode product can be expressed as $y_{i_1\cdots i_{n-1}ji_{n+1}\cdots i_{N}}=\sum_{i_n=1}^{I_n}x_{i_1i_2\cdots i_N}m_{ji_n}$. Notably, $J=1$ is equivalent to the $n$-mode product of a tensor with a vector, denoted as $\boldsymbol{\mathcal{Y}}=\boldsymbol{\mathcal{X}}\times_n \boldsymbol{m}$ where $\boldsymbol{m}\in \mathbb{R}^{I_n}$ and $\boldsymbol{\mathcal{Y}}$ is of size $I_1\times \cdots \times I_{n-1} \times I_{n+1} \times \cdots \times I_N$. Besides, we need a indicator function: $\delta_{ij}=1$ if $i=j$; otherwise $\delta_{ij}=0$.

\subsection{Spectral Graph Convolution and ChebNet}

Let $\mathcal{G}=\{\mathcal{V},\mathcal{E}\}$ be a graph with node set $\mathcal{V}$ and edge set $\mathcal{E}\subseteq \mathcal{V} \times \mathcal{V}$. 
Let $\boldsymbol{A}$ denote the adjacency matrix and $\boldsymbol{D}$ be the diagonal degree matrix. 
The normalized graph Laplacian matrix is defined as $\boldsymbol{L}=\boldsymbol{I}-\boldsymbol{D}^{-1/2}\boldsymbol{A}\boldsymbol{D}^{-1/2}$, and can be decomposed into $\boldsymbol{U}\boldsymbol{\mathit{\Lambda}}\boldsymbol{U}^\top$, where $\boldsymbol{\mathit{\Lambda}}$ is a diagonal matrix with eigenvalues of $\boldsymbol{L}$ , and $\boldsymbol{U}$ is a unitary matrix that stores the corresponding eigenvectors. 

The graph convolution between a graph signal $\boldsymbol{x}$ and a filter $\boldsymbol{g}$ is defined as $\boldsymbol{g}\ast\boldsymbol{x}=\boldsymbol{U}((\boldsymbol{U}^\top\boldsymbol{g})\odot(\boldsymbol{U}^\top\boldsymbol{x}))=\boldsymbol{U}\boldsymbol{G}\boldsymbol{U}^\top\boldsymbol{x}$. Here $\boldsymbol{U}^\top$ serves as graph Fourier transform; $\boldsymbol{G}=\operatorname{diag}(\boldsymbol{U}^\top\boldsymbol{g})$ is the graph filter in spectral domain and can be parameterized with any models. 
To avoid eigendecomposition, recent studies \cite{GCN,ChebNet} suggest that $\boldsymbol{G}$ can be approximated with a $K$-order truncated polynomial function regarding the spectrum of the normalized graph Laplacian matrix:
\begin{equation}
\label{Eq_1}
    \boldsymbol{G}\approx g_\omega(\boldsymbol{\mathit{\Lambda}})=\sum_{k=0}^{K}\omega_k\boldsymbol{\mathit{\Lambda}}^k
\end{equation}
where the parameter $\boldsymbol{\omega}\in \mathbb{R}^{K+1}$ is a vector of polynomial coefficients. Accordingly, the spectral convolution is rewritten as:
\begin{equation}
\label{Eq_2}
\boldsymbol{y}=\boldsymbol{g}\ast\boldsymbol{x}\approx\sum_{k=0}^{K}\omega_k\boldsymbol{L}^k\boldsymbol{x}
\end{equation}
where $\boldsymbol{y}$ denotes the filtered signal. 

\paragraph{ChebNet.} ChebNet \cite{ChebNet} approximates $\boldsymbol{G}$ with a truncated expansion in terms of Chebyshev polynomials:
\begin{equation}
   \label{Eq_3} \boldsymbol{y}=\boldsymbol{g}\ast\boldsymbol{x}\approx\sum_{k=0}^{K}w_kT_k(\boldsymbol{L}^\ast)\boldsymbol{x}
\end{equation}
where $\boldsymbol{L}^\ast=2\boldsymbol{L}/\lambda^\ast-\boldsymbol{I}$ denotes the scaled Laplacian matrix. Here, $\lambda^\ast$ is the largest eigenvalue of $\boldsymbol{L}$ which is not greater than $2$ and hence the eigenvalues of $\boldsymbol{L}^\ast$ lie in $[-1,1]$. $w_k$ denotes the Chebyshev coefficients. The Chebyshev polynomials can be recursively defined as $T_k(x)=2xT_{k-1}(x)-T_{k-2}(x)$ with $T_0(x)=1$ and $T_1(x)=x$. While inputting multi-channel input graph signals $\boldsymbol{X}\in\mathbb{R}^{|\mathcal{V}|\times I}$, ChebNet follows typical multi-channel CNNs that assign multiple filters to each input channel and accumulate the filtered signals in the output dimension, which we call \textbf{full coefficient setting}. Specifically, the $j$th output channel is given by:
\begin{equation}
    \label{Eq_4}
    \boldsymbol{y}_j=\sum_{k=0}^{K}\sum_{i=1}^{I}w^{k}_{ij}T_k(\boldsymbol{L}^\ast)\boldsymbol{x}_i
\end{equation}
Based on matrix multiplication rule, we can readily obtain the matrix form of \cref{Eq_4}:
\begin{equation}
    \label{Eq_5}
    \boldsymbol{Y}=\sum_{k=0}^{K}T_k(\boldsymbol{L}^\ast)\boldsymbol{X}\boldsymbol{W}_k
\end{equation}
where $w_{ij}^k$ in \cref{Eq_4} is $(i,j)$ element of the trainable weight $\boldsymbol{W}_k \in \mathbb{R}^{I\times J}$. ChebNet can learn $J$ filters for each channel of input signals at each layer, and the order-$k$ coefficient of the $j$-th filter for the $i$-th channel signal is the $(i,j)$-th element in $\boldsymbol{W}_k$. Note that $\{\boldsymbol{W}_k\}_{k=0}^{K}$ can be compiled into a coefficient tensor $\boldsymbol{\mathcal{W}}\in \mathbb{R}^{I \times J \times (K+1)}$.

\subsection{Generalizing Spectral Graph Convolution}

Although various spectral GNNs have been proposed, most of them focus on diverse graph spectrum for replacing $\boldsymbol{L}$ in \cref{Eq_2} or different polynomial bases to substitute the Chebyshev basis $T_k()$ in \cref{Eq_4}. This work does not discuss these points and without loss of generality, we set an optional graph spectrum as a sparse graph matrix $\boldsymbol{S}$ and substitutable polynomial basis as $P_k()$. For example, $\boldsymbol{S}$ can be $\boldsymbol{L}$, $\boldsymbol{L}^\ast$, normalized adjacency matrix (with or without self-loops), or other well-designed graph spectrum \cite{yang2022spectrum}; $P_k()$ can be Monomial \cite{APPNP,GPR_GNN}, Chebyshev \cite{ChebNet, ChebNetII}, Bernstein \cite{BernNet}, Jacobi \cite{JacobiConv} polynomial bases. Then, we simply extend \cref{Eq_5} to build a generalized point of view on spectral graph convolution as follows: 
\begin{equation}
    \label{Eq_6}
    \boldsymbol{Y}=\big\Vert_{j=1}^{J}\sum_{k=0}^{K}\sum_{i=1}^{I}w^{k}_{ij}P_k(\boldsymbol{S})\boldsymbol{x}_i=\sum_{k=0}^{K}P_k(\boldsymbol{S})\boldsymbol{X}\boldsymbol{W}_k
\end{equation}
where $\Vert$ is the concatenation operator. Note that we keep the full coefficient setting. Many studies \cite{SGC,APPNP,ChebNetII} generally decouple feature propagation and transformation, and thereby conclude that some methods like Vanilla GCN \cite{GCN} and APPNP \cite{APPNP} adopt fixed coefficients; and another line of methods \cite{GPR_GNN,JacobiConv,BernNet} utilize learnable coefficients. But in this work, we argue that all these SGCNs adopt learnable coefficients since feature propagation and linear transformation should be taken as an whole unified in one spectral convolution part, which can be seen as performing different tensor decomposition on the coefficient tensor $\boldsymbol{\mathcal{W}}$ under the full coefficient setting. In this point of view, the weight in linear transformation part of some SGCNs is a factor from decomposition of $\boldsymbol{\mathcal{W}}$. We will detailedly interpret this perspective by revisiting several existing SGCNs via the lens of coeffecient decomposition in \cref{UnifiedSGC}.

\section{Coefficient Decomposition for Spectral Graph Convolution}
\label{CoDeSGC}

\subsection{Unifying Existing SGCNs via lens of Coefficient Decomposition}
\label{UnifiedSGC}

In this section, we will roughly classify most of existing SGCNs into three type of architectures: multi-layer, hybrid and linear GNNs, through revisiting several typical SGCNs, and show that their convolution layers are all under the formulation in \cref{Eq_6} via lens of coefficient decomposition.

\begin{table*}[t]
\caption{A summary of $P_k()$, $\boldsymbol{S}$ and $\boldsymbol{W}_k$ in \cref{Eq_6} in typical SGCNs.}
\label{Table_method_summary}
\vskip 0.15in
\begin{center}
\begin{small}
\begin{tabular}{lcccccr}
\toprule
 & Architecture & Basis $P_k()$& Graph matrix $\boldsymbol{S}$ & Coefficient decomposition & Learnable \\
\midrule 
ChebNet & Multi-layer& Chebyshev& $\boldsymbol{L}^\ast$ & $w_{ij}^k$ & $w_{ij}^k$ \\
GCN & Multi-layer & Monomial & $\boldsymbol{\tilde{A}}$ &
$w_{ij}^k=\alpha_kw_{ij}$, 
$\alpha_k=\delta_{k1}$ & $w_{ij}$  \\
APPNP& Hybrid & Monomial & $\boldsymbol{\tilde{A}}$ & $w_{ij}^k=\alpha_kw_{ij}$, $\alpha_k=(1-\alpha)^k\alpha^{1-\delta_{kK}}$ & $w_{ij}$ \\
GPR-GNN & Hybrid & Monomial & $\boldsymbol{\tilde{A}}$ & $w_{ij}^k=\alpha_kw_{ij}$ & $w_{ij}$, $\alpha_k$  \\
ChebNetII& Hybrid & Chebyshev & $\boldsymbol{L-I}$ & $w_{ij}^k=\alpha_kw_{ij}$, $\alpha_k=\frac{2}{K+1}\sum_{l=0}^K\gamma_lT_k(x_l)$  & $w_{ij}$, $\gamma_l$  \\
BernNet & Hybrid & Bernstein & $\boldsymbol{L}/2$ & $w_{ij}^k=\alpha_kw_{ij}$ & $w_{ij}$, $\alpha_k$\\
JacobiConv & Linear & Jacobi & $\boldsymbol{I-L}$ & $w_{ij}^k=\alpha_{kj}w_{ij}$, $\alpha_{kj}=\beta_{kj}\prod_{l=1}^{k}\gamma^\prime\operatorname{tanh}\eta_l$ & $w_{ij}$, $\beta_{kj}$, $\eta_l$ \\
FavardGNN & Hybrid & Favard & $\boldsymbol{I-L}$ & $w_{ij}^k=\alpha_{ki}w_{ij}$ & $w_{ij}$, $\alpha_{ki}$\\
\bottomrule
\end{tabular}
\end{small}
\end{center}
\vskip -0.1in
\end{table*}

\paragraph{Vanilla GCN.} The vanilla GCN \cite{GCN} sets $K=1$, $\lambda^\ast=2$, $w_{ij}^0=-w_{ij}^1=w_{ij}$ in \cref{Eq_4}, and then obtains the convolution layer as follows:
\begin{equation}
    \label{Eq_7}
    \begin{aligned}
    \boldsymbol{y}_j=\sum_{i=1}^{I}w_{ij}\boldsymbol{x}_i-\sum_{i=1}^{I}w_{ij}(\boldsymbol{L-I})\boldsymbol{x}_i\\
    =\sum_{i=1}^{I}w_{ij}(\boldsymbol{I}+\boldsymbol{D}^{-1/2}\boldsymbol{A}\boldsymbol{D}^{-1/2})\boldsymbol{x}_i
    \end{aligned}
\end{equation}
Through the renormalization trick that sets $\boldsymbol{I}+\boldsymbol{D}^{-1/2}\boldsymbol{A}\boldsymbol{D}^{-1/2}\approx\boldsymbol{\hat{D}}\,\!^{-1/2}\boldsymbol{\hat{A}}\boldsymbol{\hat{D}}\,\!^{-1/2}=\boldsymbol{\tilde{A}}$, where $\boldsymbol{\hat{A}=A+I}$, $\boldsymbol{\hat{D}=D+I}$, the matrix form of \cref{Eq_7} can be written as:
\begin{equation}
    \label{Eq_8}   
    \boldsymbol{Y}=(\boldsymbol{I}+\boldsymbol{D}^{-1/2}\boldsymbol{A}\boldsymbol{D}^{-1/2})\boldsymbol{X}\boldsymbol{W}\approx\boldsymbol{\tilde{A}}\boldsymbol{X}\boldsymbol{W}
\end{equation}
Vanilla GCN keeps the same architecture as ChebNet, where multiple convolution layers are connected by nonlinear activation. We call them \textbf{multi-layer} GNNs. Obviously, each layer of vanilla GCN (\cref{Eq_8}) is under the formulation \cref{Eq_6} by setting $\boldsymbol{S=\tilde{A}}$ and $P_k(\lambda)=\lambda^k$, and performing a coefficient decomposition $w_{ij}^k=\alpha_kw_{ij}$ while fixing $\alpha_k=\delta_{k1}$.
Essentially, vanilla GCN defines its convolution layer by the aforementioned coefficient decomposition fixing the linear relationship among the frontal slices in the coefficient tensor $\boldsymbol{\mathcal{W}}$. 

\paragraph{GPR-GNN.} GPR-GNN \cite{GPR_GNN} first extracts hidden state features for each node and then uses Generalized PageRank (GPR) to propagate them. Given the initialized node features $\boldsymbol{F}\in \mathbb{R}^{|\mathcal{V}| \times F}$, the GPR-GNN model architecture can be described as:
\begin{equation}
    \label{Eq_9}
    \boldsymbol{Y}=\sum_{k=0}^{K}\alpha_k\boldsymbol{\tilde{A}}^kf_{\theta}(\boldsymbol{F})
\end{equation}
where $f_{\theta}(\cdot)$ is a neural network with parameter set ${\theta}$, usually implementing a 2-layer MLP and the GPR weights $\alpha_k$ is trainable. Scrutinizing the implementation, we find there is no any nonlinear activation following the final linear layer of $f_{\theta}(\cdot)$. Accordingly, the final linear layer should be taken together with the GPR propagation as an whole graph convolution layer. By ignoring the bias in the final linear layer (we will discuss the role of bias in the follows), we can rewrite \cref{Eq_9} as follows:
\begin{equation}
    \label{Eq_10}
    \boldsymbol{Y}=\sum_{k=0}^{K}\alpha_k\boldsymbol{\tilde{A}}^k(f^\prime_{\theta^\prime}(\boldsymbol{F})\boldsymbol{W})=\sum_{k=0}^{K}\boldsymbol{\tilde{A}}^k\boldsymbol{X}\alpha_k\boldsymbol{W}
\end{equation}
where $f^\prime_{\theta^\prime}(\cdot)$ is the nonlinear part in $f_{\theta}(\cdot)$ and $\boldsymbol{X}=f^\prime_{\theta^\prime}(\boldsymbol{F})$ is the input graph signals. Obviously, the whole GPR-GNN architecture consists of a nonlinear feature transformation part and a linear graph convolution part. We call SGCNs under this architecture \textbf{hybrid} GNNs, such as APPNP \cite{APPNP}, BernNet \cite{BernNet}, ChebNetII \cite{ChebNetII}, FavardGNN/OptBasisGNN \cite{Optbasis} and so on. We can also observe that the convolution part \cref{Eq_10} is under the formulation \cref{Eq_6} with a coefficient decomposition $\boldsymbol{W}_k=\alpha_k\boldsymbol{W}$, $\boldsymbol{S}=\boldsymbol{\tilde{A}}$ and Monomial basis $P_k(\lambda)=\lambda^k$. Actually, vanilla GCN and GPR-GNN use the same approach to decomposing the coefficient tensor, while the former fixes and the latter learns the linear relationship $\alpha_k$ across the frontal slices of the coefficient tensor.  

\paragraph{JacobiConv.} 
JacobiConv \cite{JacobiConv} first feeds node features $\boldsymbol{F}$ into a linear layer with bias $\boldsymbol{\hat{X}}=\boldsymbol{FW}+\boldsymbol{1b^\top}$ and filters $\boldsymbol{\hat{X}}$ by:
\begin{equation}
    \label{Eq_11}
    \boldsymbol{Y}=\big\Vert_{j=1}^{J}\sum_{k=0}^{K}\alpha_{kj}P_k(\boldsymbol{I-L})\boldsymbol{\hat{x}}_j
\end{equation}
where $P_k(\cdot)$ is set as the Jacobi basis whose form is provided in \cref{JacobiBasis}. Note that JacobiConv deserts nonlinearity, we call it a \textbf{linear} GNN.  We find that if we treat node features $\boldsymbol{F}$ as the input signals $\boldsymbol{X}$, and we have $\boldsymbol{\hat{x}}_j=\sum_{i=1}^{I}w_{ij}\boldsymbol{x}_i+b_j\boldsymbol{1}$, then \cref{Eq_11} can be rewritten as:
\begin{equation}
    \label{Eq_12}
    \boldsymbol{Y}=\big\Vert_{j=1}^{J}\sum_{k=0}^{K}\sum_{i=0}^{I}\alpha_{kj}w_{ij}P_k(\boldsymbol{I-L})\boldsymbol{x}_i
\end{equation}
where $w_{0j}=b_j$ and  $\boldsymbol{x}_0=\boldsymbol{1}$. Obviously, using the bias in linear transformation does not change the architecture of the linear GNN and is equivalent to adding an extra channel of signal $\boldsymbol{x}_0=\boldsymbol{1}$. Interestingly, we readily find the facts that \cref{Eq_12} is still under the formulation \cref{Eq_6} with $\boldsymbol{S=I-L}$, choosing Jacobi basis, and a decomposition on coefficient tensor $w_{ij}^k=\alpha_{kj}w_{ij}$. It is worth mentioning that JacobiConv further designs a so-called polynomial coefficient decomposition technique that decomposes $\alpha_{kj}$ into $\beta_{kj}\prod_{l=1}^{k}\gamma_{l}$ and sets $\gamma_l=\gamma^\prime\operatorname{tanh}\eta_l$ to enforce $\gamma_l\in [-\gamma^\prime, \gamma^\prime]$. In essence, this trick only substitutes the whole coefficient decomposition process with $w_{ij}^k=w_{ij}\beta_{kj}\prod_{l=1}^{k}\gamma_l$. Accordingly, the whole JacobiConv architecture can be viewed as a single graph convolution layer under the formulation \cref{Eq_6}. 

To sum up, each layer of multi-layer GNNs, the convolution part of hybrid GNNs and the whole of a linear GNN are all spectral graph convolutions under the formulation \cref{Eq_6} with different choices for graph matrix $\boldsymbol{S}$, polynomial basis $P_k$ and coefficient decomposition on the tensor $\boldsymbol{\mathcal{W}}$. Under the general form of spectral graph convolution given in \cref{Eq_6}, several typical spectral GNNs are summarized in \cref{Table_method_summary} and \cref{OtherHybrid}. It is worth mentioning that since there exist learnable factor parameters in each model, all of them can be supposed to be able to learn filters. Inspired by this perspective, we derive two novel spectral graph convolution architectures through CP and Tucker decomposition which will be shown in \cref{CoDeSGC-CPAndTucker}.

\subsection{Coefficient Decomposition with CP and Tucker for Spectral Graph Convolution}
\label{CoDeSGC-CPAndTucker}

As mentioned above, vanilla GCN substantially fixes linear relationship among the coefficients acorss diffrent orders in each filter and GPR-GNN can adaptively learn the linear relations. Both of them share the same linear relationship for all $I\times J$ filters (each input channel has $J$ filters). By contrast, as shown in \cref{Eq_12}, JacobiConv realizes different linear relations in the component filters for different output channels. Motivated by this insight, we consider tensor decomposition like CP and Tucker decomposition to extend these convolution operators and derive novel architectures that can learn more sophisticated multilinear relations between the polynomial coefficients stored in the tensor $\boldsymbol{\mathcal{W}}$.  

\paragraph{CoDeSGC-CP.} We refer to CANDECOMP/PARAFAC decomposition as CP \cite{Tensor_decomposition,multiway}. In a three-mode case, CP decomposes a third-order tensor into three matrices with the same column size. Specifically, given a third-order coefficient tensor $\boldsymbol{\mathcal{W}}\in \mathbb{B}^{I \times J \times (K+1)}$ where $I$, $J$ and $K$ are respectively the number of input channels, the number of output channels and the order of truncated polynomial filter, we have its CP decomposition as follows:
\begin{equation}
    \label{Eq_13}
    \boldsymbol{\mathcal{W}}=\sum_{r=1}^{R}\boldsymbol{c}_r\circ\boldsymbol{p}_r\circ\boldsymbol{m}_r=[\![\boldsymbol{C}, \boldsymbol{P}, \boldsymbol{M}]\!]
\end{equation}
where $\boldsymbol{C} \in \mathbb{R}^{I \times R}$, $\boldsymbol{P} \in \mathbb{R}^{J \times R}$ and $\boldsymbol{M} \in \mathbb{R}^{(K+1) \times R}$ are the factor matrices; the $r$th column is respectively $\boldsymbol{c}_r$, $\boldsymbol{p}_r$ and $\boldsymbol{m}_r$. Recall that $\circ$ is the outer product of vectors, the element-wise formulation is written as $w^k_{ij}=\sum_{r=1}^{R}c_{ir}p_{jr}m_{kr}$. In terms of the frontal slices of $\boldsymbol{\mathcal{W}}$, CP decomposition can also be rewritten as $\boldsymbol{W}_k=\boldsymbol{C}\operatorname{diag}(\boldsymbol{m}_{k:})\boldsymbol{P}^\top$. Therefore, our CoDeSGC-CP layer can be formulated by:
\begin{equation}
    \label{Eq_14}
    \begin{aligned}
        \boldsymbol{Y}=\big\Vert_{j=1}^{J}\sum_{k=0}^{K}\sum_{i=1}^{I}\sum_{r=1}^{R}c_{ir}p_{jr}m_{kr}P_k(\boldsymbol{S})\boldsymbol{x}_i\\=\sum_{k=0}^{K}P_k(\boldsymbol{S})\boldsymbol{X}\boldsymbol{C}\operatorname{diag}(\boldsymbol{m}_{k:})\boldsymbol{P}^\top
    \end{aligned}
\end{equation}
For the effectiveness of computation, we can first compute $\boldsymbol{H}=\boldsymbol{XC}$ to reduce the dimensionality of the input signals, followed by the propagation process $\boldsymbol{Z}=\sum_{k=0}^{K}P_k(\boldsymbol{S})\boldsymbol{H}\operatorname{diag(\boldsymbol{m}_{k:})}$ which usually can be computed by iteratively exploiting sparse matrix multiplication. Without loss of generality, we set $\boldsymbol{V}_k=P_k(\boldsymbol{S})\boldsymbol{H}$ and simply denote the each iteration of the propagation process as $\boldsymbol{V}_k=\Psi\left(\{\boldsymbol{V}_i\}_{i=0}^{k-1},\boldsymbol{S}\right)$, then we obtain $\boldsymbol{Z}$ by $\sum_{k=0}^{K}\boldsymbol{V}_k\operatorname{diag}(\boldsymbol{m}_{k:})$. Finally, we compute a linear transform to obtain the output signals $\boldsymbol{Y}=\boldsymbol{ZP}^\top$. We summarize the spectral convolution process in \cref{alg_CP}, where the bias is used in each linear transformation step.  

\begin{algorithm}[t]
   \caption{CoDeSGC-CP}
   \label{alg_CP}
   {\bfseries Input:} Input signals $\boldsymbol{X} \in \mathbb{R}^{|\mathcal{V}|\times I}$; Sparse graph matrix $\boldsymbol{S}$; Truncated polynomial order $K$\\
   {\bfseries Learnable Parameters:} $\boldsymbol{C} \in \mathbb{R}^{I \times R}$, $\boldsymbol{b}_{C} \in \mathbb{R}^{R}$,   $\boldsymbol{P} \in \mathbb{R}^{J \times R}$, $\boldsymbol{b}_{P} \in \mathbb{R}^J$,  $\boldsymbol{M} \in \mathbb{R}^{(K+1) \times R}$\\
   {\bfseries Output:} Output signals $\boldsymbol{Y}\in \mathbb{R}^{|\mathcal{V}|\times J}$
\begin{algorithmic}[1]
   \STATE $\boldsymbol{H} \gets  \boldsymbol{XC}+\boldsymbol{1}\boldsymbol{b}^\top_{C}$ 
   \STATE $\boldsymbol{Z} \gets \sum_{k=0}^{K}P_k(\boldsymbol{S})\boldsymbol{H}\operatorname{diag}(\boldsymbol{m}_{k:})$ by steps 3-6: 
   \STATE  \qquad $\boldsymbol{V}_0 \gets P_0(\boldsymbol{S})\boldsymbol{H}$, $\boldsymbol{Z}\gets \boldsymbol{V}_0\operatorname{diag}(\boldsymbol{m}_{0:})$ 
   \STATE \qquad \textbf{for} $k=1$ \textbf{to} $K$ \textbf{do}
   \STATE \qquad \quad \ $\boldsymbol{V}_k \gets \Psi\left(\{\boldsymbol{V}_i\}_{i=0}^{k-1},\boldsymbol{S}\right)$ \\ \qquad  \textcolor{blue}{/\,/ \ iteratively perform sparse matrix multiplication}
   \STATE \qquad \quad \ $\boldsymbol{Z}\gets \boldsymbol{Z}+\boldsymbol{V}_k\operatorname{diag}(\boldsymbol{m}_{k:})$ 
   \STATE $\boldsymbol{Y} \gets \boldsymbol{Z}\boldsymbol{P}^\top+\boldsymbol{1}\boldsymbol{b}^\top_P$
   
\end{algorithmic}
{\bfseries return:} $\boldsymbol{Y}$
\end{algorithm}

Evidently, the decomposition in \cref{Eq_12}, i.e., $w^k_{ij}=\alpha_{kj}w_{ij}$ is a specific reduced version of the CP decomposition, as $m_{kr}$ and $c_{ir}$ become equivalent to $\alpha_{kj}$ and $w_{ij}$ while setting $R=J$ and $p_{jr}=\delta_{jr}$ in \cref{Eq_14}. We naturally can adopt the similar trick as JacobiConv to further decompose $m_{kr}$, i.e., $m_{kr}=\beta_{kr}\prod_{l=1}^{k}\gamma_l$ and $\gamma_l=\gamma^\prime\operatorname{tanh}\eta_l$, or slightly more complex decomposition, e.g., $m_{kr}=\beta_{kr}\prod_{l=1}^{k}\gamma_{lr}$. But, empirically according to JacobiConv  \cite{JacobiConv}, imposing such trick can only marginally improve the performance in some cases and even worse the performance in minority cases. Moreover, it also brings about additional burden on model training and selection due to the addition of hyperparameters and trainable parameters (hyperparameter $\gamma^\prime$ as well as the different learning rates and weight decay respectively for trainable $\beta_{kr}$ and $\eta_{l}$ as JacobiConv implements). Thereby, we think this trick seems not to be essential in our model, since CoDeSGC-CP provides sufficient performance gain which will be shown in \cref{results}.


\paragraph{CoDeSGC-Tucker.} The Tucker decomposition decomposes a tensor into a core tensor multiplied by a matrix along each dimension \cite{Tensor_decomposition}, which implies more intricate multiplex linear relations than CP. Formally, given a third-order coefficient tensor $\boldsymbol{\mathcal{W}}$, we have its Tucker decomposition as follows:
\begin{equation}
    \label{Eq_15}
    \begin{aligned}
        \boldsymbol{\mathcal{W}}&=\boldsymbol{\mathcal{G}}\times_1 \boldsymbol{C}\times_2\boldsymbol{P}\times_3\boldsymbol{M}\\&=\sum_{p=1}^P\sum_{q=1}^Q\sum_{r=1}^{R}g_{pqr}\boldsymbol{c}_p\circ\boldsymbol{p}_q\circ\boldsymbol{m}_r=[\![\boldsymbol{\mathcal{G}};\boldsymbol{C}, \boldsymbol{P}, \boldsymbol{M}]\!]
    \end{aligned}
\end{equation}
where $\boldsymbol{c}_p$, $\boldsymbol{p}_q$ and $\boldsymbol{m}_r$ are the $p$th, $q$th and $r$th column of the factor matrices $\boldsymbol{C} \in \mathbb{R}^{I \times P}$, $\boldsymbol{P} \in \mathbb{R}^{J \times Q}$ and $\boldsymbol{M} \in \mathbb{R}^{(K+1) \times R}$; the tensor $\boldsymbol{\mathcal{G}}\in \mathbb{R}^{P\times Q\times R}$ is called core tensor which stores the weights for component tensors $\boldsymbol{c}_p\circ\boldsymbol{p}_q\circ\boldsymbol{m}_r$. Obviously, while setting $P=Q=R$ and $g_{pqr}=\delta_{pq}\delta_{qr}$, the Tucker decomposition is reduced to CP decomposition. Combining the element-wise form of Tucker decomposition of coeffecient tensor with \cref{Eq_6}, we can write the CoDeSGC-Tucker as follows:
\begin{equation}
    \label{Eq_16}
    \boldsymbol{Y}=\big\Vert_{j=1}^{J}\sum_{k=0}^{K}\sum_{i=1}^{I}\sum_{p=1}^P\sum_{q=1}^Q\sum_{r=1}^{R}g_{pqr}c_{ip}p_{jq}m_{kr}P_k(\boldsymbol{S})\boldsymbol{x}_i
\end{equation}
Note that $\sum_{i=1}^{I}\sum_{p=1}^{P}g_{pqr}c_{ip}\boldsymbol{x}_i$ is the $(q, r)$th mode-1 fiber $\boldsymbol{h}_{:qr}$ of $\boldsymbol{\mathcal{H}}=\boldsymbol{\mathcal{G}}\times_1(\boldsymbol{XC})$. Also, the 1-mode product can be expressed in terms of mode-1 unfolding: $\boldsymbol{H}_{(1)}=(\boldsymbol{XC})\boldsymbol{G}_{(1)}$. Therefore, we can utilize linear transformations $\boldsymbol{C}$ and $\boldsymbol{G}_{(1)}$ to convert input signals $\boldsymbol{X}$ into $\boldsymbol{H}_{(1)}$. Then, \cref{Eq_16} is simplified as:
\begin{equation}
    \label{Eq_17}    \boldsymbol{Y}=\big\Vert_{j=1}^{J}\sum_{k=0}^{K}\sum_{q=1}^Q\sum_{r=1}^{R}p_{jq}m_{kr}P_k(\boldsymbol{S})\boldsymbol{h}_{:qr}
\end{equation}
Subsequently, we maintain the similar computation process as CoDeSGC-CP: propagation then transformation. To be more specific, \cref{Eq_17} can be rewritten as:
\begin{equation}
    \label{Eq_18}    \boldsymbol{Y}=\left(\big\Vert_{q=1}^{Q}\sum_{k=0}^{K}\sum_{r=1}^{R}m_{kr}P_k(\boldsymbol{S})\boldsymbol{h}_{:qr}\right)\boldsymbol{P}^\top
\end{equation}
So-called propagation then transformation refers to computing $\boldsymbol{Z}=\big\Vert_{q=1}^{Q}\sum_{k=0}^{K}\sum_{r=1}^{R}m_{kr}P_k(\boldsymbol{S})\boldsymbol{h}_{:qr}$ and then acquiring the output signals $\boldsymbol{Y}$ by $\boldsymbol{Z}\boldsymbol{P}^\top$. 

\begin{algorithm}[t]
   \caption{CoDeSGC-Tucker}
   \label{alg_Tucker}
   {\bfseries Input:} Input signals $\boldsymbol{X} \in \mathbb{R}^{|\mathcal{V}|\times I}$; Sparse graph matrix $\boldsymbol{S}$; Truncated polynomial order $K$\\
   {\bfseries Learnable Parameters:} $\boldsymbol{C} \in \mathbb{R}^{I \times P}$, $\boldsymbol{b}_{C} \in \mathbb{R}^{P}$, $\boldsymbol{P} \in \mathbb{R}^{J \times Q}$, $\boldsymbol{b}_{P} \in \mathbb{R}^Q$, $\boldsymbol{G}_{(1)}\in \mathbb{R}^{P\times QR}$, $\boldsymbol{b}_{G}\in \mathbb{R}^{QR}$,   $\boldsymbol{M} \in \mathbb{R}^{(K+1) \times R}$\\
   {\bfseries Output:} Output signals $\boldsymbol{Y}\in \mathbb{R}^{|\mathcal{V}|\times J}$
\begin{algorithmic}[1]
   \STATE $\boldsymbol{H}_{(1)} \gets  (\boldsymbol{XC}+\boldsymbol{1}\boldsymbol{b}^\top_{C})\boldsymbol{G}_{(1)}+\boldsymbol{1}\boldsymbol{b}^\top_{G}$;  
   \STATE $\boldsymbol{Z} \gets\big\Vert_{q=1}^{Q}\sum_{k=0}^{K}\sum_{r=1}^{R}m_{kr}P_k(\boldsymbol{S})\boldsymbol{h}_{:qr}$ by steps 3-7: 
   \STATE \qquad $(\boldsymbol{V}_0)_{(1)} \gets P_0(\boldsymbol{S})\boldsymbol{H}_{(1)}$,
   \\ \qquad $\boldsymbol{\mathcal{V}}_0 \gets \operatorname{reshape}((\boldsymbol{V}_0)_{(1)},(|\mathcal{V}|,Q,R))$,
   \\ \qquad $\boldsymbol{Z}\gets \boldsymbol{\mathcal{V}}_0 \times_3 \boldsymbol{m}_{0:}$
   \STATE \qquad \textbf{for} $k=1$ \textbf{to} $K$ \textbf{do}
   \STATE \qquad \quad \ $(\boldsymbol{V}_k)_{(1)} \gets \Psi\left(\{(\boldsymbol{V}_i)_{(1)}\}_{i=0}^{k-1},\boldsymbol{S}\right)$ \\  \qquad \textcolor{blue}{/\,/ \ iteratively perform sparse matrix multiplication}
   \STATE \qquad \quad \ $\boldsymbol{\mathcal{V}}_k \gets \operatorname{reshape}((\boldsymbol{V}_k)_{(1)},(|\mathcal{V}|,Q,R))$
   \STATE \qquad \quad \ $\boldsymbol{Z}\gets \boldsymbol{Z}+\boldsymbol{\mathcal{V}}_k\times_3\boldsymbol{m}_{k:}$ 
   \STATE $\boldsymbol{Y} \gets \boldsymbol{Z}\boldsymbol{P}^\top+\boldsymbol{1}\boldsymbol{b}^\top_P$
   
\end{algorithmic}
{\bfseries return:} $\boldsymbol{Y}$
\end{algorithm}

For the propagation process, notice that $P_k(\boldsymbol{S})\boldsymbol{h}_{:qr}$ is the $(q,r)$th mode-1 fiber $\boldsymbol{v}^k_{:qr}$ of tensor $\boldsymbol{\mathcal{V}}_k=\boldsymbol{\mathcal{H}}\times_1 P_k(\boldsymbol{S})$. To make full use of sparse matrix multiplication, we compute $\boldsymbol{\mathcal{V}}_k$ through two step: we first obtain its mode-1 unfolding by $(\boldsymbol{V}_k)_{(1)}=P_k(\boldsymbol{S})\boldsymbol{H}_{(1)}$ and then utilize $\operatorname{reshape}()$ operator to return it. The first step can be completed via the similar iterative approach as CoDeSGC-CP and without loss of generality, we denote the each iteration as $(\boldsymbol{V}_k)_{(1)}= \Psi\left(\{(\boldsymbol{V}_i)_{(1)}\}_{i=0}^{k-1},\boldsymbol{S}\right)$. After obtaining $\boldsymbol{\mathcal{V}}_k$, the propagation process can be reduced to $\boldsymbol{Z}=\sum_{k=0}^{K}(\big\Vert_{q=1}^{Q}\sum_{r=1}^{R}m_{kr}\boldsymbol{v}^k_{:qr})$; on the other hand, note that $\big\Vert_{q=1}^{Q}\sum_{r=1}^{R}m_{kr}\boldsymbol{v}^k_{:qr}=\boldsymbol{\mathcal{V}}_k \times_3 \boldsymbol{m}_{k:}$, we can obtain $\boldsymbol{Z}$ by the iterative summation. Similarly, we summarize the forward process of CoDeSGC-Tucker in \cref{alg_Tucker}, where the bias is used in each linear transformation step.




\section{Experiment}
\label{results}

\subsection{Experimental Settings}

\paragraph{Datasets.} 
To provide a reasonable evaluation of our methods, we consider several graph datasets conventionally used in preceding graph filtering works, including 5 homophilic datasets: citation graphs Cora, CiteSeer, PubMed \cite{coraet}, and co-purchase graphs Computers and Photo \cite{fourdata}; 5 heterophilic datasets: Wikipedia graphs  Chameleon and Squirrel \cite{musae}, the Actor cooccurrence graph and webpage graphs Texas and Cornell from WebKB \cite{Pei2020Geom}. The statistics of these datasets are summarized in \cref{ExperimentSetUp} \cref{Table_datasets}.

\paragraph{Baselines.} We mainly test our proposed methods in linear GNNs like JacobiConv \cite{JacobiConv}. To be more specific, for the node classification task, we take the node features as the input signals, and obtain the filtered signals which are then directly fed into Softmax to get the predictions of labels for nodes. Unless specific statement, we use the same settings for polynomial basis and graph matrix as JacobiConv, i.e., Jacobi basis and normalized adjacency matrix. As shown in \cref{alg_CP} and \cref{alg_Tucker}, the bias is used in each linear transformation. We also consider other spectral GNNs as baselines including GCN, APPNP, ChebNet, GPRGNN, BernNet and JacobiConv.      

\begin{table*}[t]
\caption{Results on real-world datasets: Mean accuracy ($\%$) $\pm 95\%$ confidence interval.}
\label{Table_results}
\vskip 0.15in
\setlength\tabcolsep{4.5pt}
\begin{center}
\begin{small}
\begin{tabular}{lcc|ccc|ccc}
\hline
 Datasets& ChebNet & GCN & APPNP & GPR-GNN & BernNet & JacobiConv & CoDeSGC-Tucker& CoDeSGC-CP \\
  
\hline 
Cora& $86.67$\tiny$\pm0.82$&$87.14$\tiny$\pm1.01$&$88.14$\tiny$\pm0.73$&$88.57$\tiny$\pm0.69$&$88.52$\tiny$\pm0.95$ &$88.98$\tiny$\pm0.46$& $\mathbf{89.41}$\tiny$\mathbf{\pm0.85}$& $89.23$\tiny$\pm0.71$   \\
CiteSeer& $79.11$\tiny$\pm0.75$&$79.86$\tiny$\pm0.67$&$80.47$\tiny$\pm0.74$&$80.12$\tiny$\pm0.83$&$80.09$\tiny$\pm0.79$ &$80.78$\tiny$\pm0.79$&$81.26$\tiny$\pm0.55$ &$\mathbf{81.31}$\tiny$\mathbf{\pm0.91}$  \\
PubMed& $87.95$\tiny$\pm0.28$&$86.74$\tiny$\pm0.27$&$88.12$\tiny$\pm0.31$&$88.46$\tiny$\pm0.33$&$88.48$\tiny$\pm0.41$ &$89.62$\tiny$\pm0.41$& $\mathbf{91.26}$\tiny$\mathbf{\pm0.35}$& $91.25$\tiny$\pm0.48$ \\
Computers & $87.54$\tiny$\pm0.43$&$83.32$\tiny$\pm0.33$&$85.32$\tiny$\pm0.37$&$86.85$\tiny$\pm0.25$&$87.64$\tiny$\pm0.44$ &$90.39$\tiny$\pm0.29$& $\mathbf{91.31}$\tiny$\mathbf{\pm0.31}$ & $91.06$\tiny$\pm0.17$ \\
Photo& $93.77$\tiny$\pm0.32$&$88.26$\tiny$\pm0.73$&$88.51$\tiny$\pm0.31$&$93.85$\tiny$\pm0.28$&$93.63$\tiny$\pm0.35$ &$95.43$\tiny$\pm0.23$& $\mathbf{95.61}$\tiny$\mathbf{\pm0.29}$&$95.54$\tiny$\pm0.28$\\
Chameleon & $59.28$\tiny$\pm1.25$&$59.61$\tiny$\pm2.21$&$51.84$\tiny$\pm1.82$&$67.28$\tiny$\pm1.09$&$68.29$\tiny$\pm1.58$ &$74.20$\tiny$\pm1.03$&$\mathbf{75.56}$\tiny$\mathbf{\pm0.98}$&$74.53$\tiny$\pm1.14$  \\
Actor& $37.61$\tiny$\pm0.89$&$33.23$\tiny$\pm1.16$&$39.66$\tiny$\pm0.55$&$39.92$\tiny$\pm0.67$&$41.79$\tiny$\pm1.01$ &$41.17$\tiny$\pm0.64$& $\mathbf{43.13}$\tiny$\mathbf{\pm0.96}$&$41.98$\tiny$\pm1.00$\\
Squirrel& $40.55$\tiny$\pm0.42$&$46.78$\tiny$\pm0.87$&$34.71$\tiny$\pm0.57$&$50.15$\tiny$\pm1.92$&$51.35$\tiny$\pm0.73$ &$57.38$\tiny$\pm1.25$&$\mathbf{63.78}$\tiny$\mathbf{\pm0.90}$ &$62.68$\tiny$\pm0.59$\\
Texas &$86.22$\tiny$\pm2.45$&$77.38$\tiny$\pm3.28$&$90.98$\tiny$\pm1.64$&$92.95$\tiny$\pm1.31$&$93.12$\tiny$\pm0.65$ &$\mathbf{93.44}$\tiny$\mathbf{\pm2.13}$&$91.15$\tiny$\pm2.30$&$90.66$\tiny$\pm2.62$\\
Cornell& $83.93$\tiny$\pm2.13$&$65.90$\tiny$\pm4.43$&$91.81$\tiny$\pm1.96$&$91.37$\tiny$\pm1.81$&$92.13$\tiny$\pm1.64$ &$\mathbf{92.95}$\tiny$\mathbf{\pm2.46}$&$90.64$\tiny$\pm2.55$&$89.57$\tiny$\pm2.98$\\
\hline
\end{tabular}
\end{small}
\end{center}
\vskip -0.1in
\end{table*}

\begin{table*}[t]
\caption{Results of ablation study on real-world datasets: Mean accuracy ($\%$) $\pm 95\%$ confidence interval.}
\label{Table_ablation}
\vskip 0.15in
\setlength\tabcolsep{5pt}
\begin{center}
\begin{small}
\begin{tabular}{lcccc|cccc}
\hline
 Datasets& H-Tucker2 & H-Tucker1 & H-Tucker & H-CP & L-Tucker2 & L-Tucker1 & L-Tucker & L-CP  \\
  
\hline 
Cora& $88.65$\tiny$\pm0.94$&$88.80$\tiny$\pm0.95$&$\mathbf {89.39}$\tiny$\mathbf{\pm0.94}$&$88.72$\tiny$\pm0.77$&$89.10$\tiny$\pm0.85$ &$89.00$\tiny$\pm0.84$& $\mathbf{89.41}$\tiny$\mathbf{\pm0.85}$& $89.23$\tiny$\pm0.71$  \\
CiteSeer& $80.38$\tiny$\pm0.55$&$\mathbf{81.58}$\tiny$\mathbf{\pm0.90}$&$80.35$\tiny$\pm0.97$&$80.30$\tiny$\pm0.64$& $80.82$\tiny$\pm0.83$&$80.03$\tiny$\pm0.90$&$81.26$\tiny$\pm0.55$ &$\mathbf{81.31}$\tiny$\mathbf{\pm0.91}$  \\
PubMed& $91.61$\tiny$\pm0.29$&$91.60$\tiny$\pm0.27$&$\mathbf{91.84}$\tiny$\mathbf{\pm0.33}$&$91.51$\tiny$\pm0.35$&$\mathbf{91.45}$\tiny$\mathbf{\pm0.54}$ &$90.78$\tiny$\pm0.42$& $91.26$\tiny$\pm0.35$& $91.25$\tiny$\pm0.48$\\
Computers & $\mathbf{91.95}$\tiny$\mathbf{\pm0.18}$&$91.29$\tiny$\pm0.22$&$91.48$\tiny$\pm0.23$&$91.83$\tiny$\pm0.19$&$91.15$\tiny$\pm0.17$ &$90.51$\tiny$\pm0.18$& $\mathbf{91.31}$\tiny$\mathbf{\pm0.31}$ & $91.06$\tiny$\pm0.17$ \\
Photo& $\mathbf{95.95}$\tiny$\mathbf{\pm0.28}$&$95.87$\tiny$\pm0.24$&$95.12$\tiny$\pm0.20$&$95.19$\tiny$\pm0.34$&$95.30$\tiny$\pm0.16$ &$\mathbf{95.86}$\tiny$\mathbf{\pm0.27}$& $95.61$\tiny$\pm0.29$&$95.54$\tiny$\pm0.28$\\
Chameleon & $74.44$\tiny$\pm0.63$&$\mathbf{74.79}$\tiny$\mathbf{\pm0.94}$&$73.83$\tiny$\pm1.20$&$73.92$\tiny$\pm1.14$&$74.79$\tiny$\pm0.85$&$74.40$\tiny$\pm0.92$ &$\mathbf{75.56}$\tiny$\mathbf{\pm0.98}$&$74.53$\tiny$\pm1.14$  \\
Actor& $42.25$\tiny$\pm0.78$&$42.51$\tiny$\pm0.77$&$42.24$\tiny$\pm0.66$&$\mathbf{43.89}$\tiny$\mathbf{\pm0.98}$&$41.89$\tiny$\pm0.98$&$39.85$\tiny$\pm0.67$& $\mathbf{43.13}$\tiny$\mathbf{\pm0.96}$&$41.98$\tiny$\pm1.00$ \\
Squirrel& $64.43$\tiny$\pm0.70$&$64.06$\tiny$\pm1.13$&$64.44$\tiny$\pm0.99$&$\mathbf{64.69}$\tiny$\mathbf{\pm0.94}$&$63.53$\tiny$\pm0.92$ &$63.25$\tiny$\pm0.72$&$\mathbf{63.78}$\tiny$\mathbf{\pm0.90}$ &$62.68$\tiny$\pm0.59$\\
Texas &$87.70$\tiny$\pm4.92$&$\mathbf{90.66}$\tiny$\mathbf{\pm1.64}$&$88.85$\tiny$\pm3.77$&$86.56$\tiny$\pm2.95$&$86.07$\tiny$\pm4.27$ &$88.85$\tiny$\pm4.10$&$\mathbf{91.15}$\tiny$\mathbf{\pm2.30}$&$90.66$\tiny$\pm2.62$\\
Cornell& $89.79$\tiny$\pm2.98$&$\mathbf{91.06}$\tiny$\mathbf{\pm3.83}$&$85.32$\tiny$\pm2.77$&$86.60$\tiny$\pm6.18$&$86.38$\tiny$\pm3.19$ &$82.98$\tiny$\pm5.96$&$\mathbf{90.64}$\tiny$\mathbf{\pm2.55}$&$89.57$\tiny$\pm2.98$\\
\hline
\end{tabular}
\end{small}
\end{center}
\vskip -0.1in
\end{table*}

\paragraph{Experimental Setup.} We follow the same evaluation protocol as JacobiConv \cite{JacobiConv}. Specifically, the node set in each dataset is randomly split into train/validation/test sets with a ratio of $60\%$/$20\%$/$20\%$. We run repeating experiments with different random initialization and random data splits. Concretely, the average performances on validation sets in $3$ runs are used to select hyperparameters of our models by Optuna \cite{akiba2019optuna} for $400$ trails, and we report the average performances with a $95\%$ confidence interval on test sets in $10$ runs with the best hyperparameters, which exactly accords with JacobiConv. Also following \cite{JacobiConv}, we fix the order of polynomial $K=10$, and train models in $1000$ epochs using early-stopping strategy with a patience of 200 epochs. More configuration details about model hyperparameters and computing infrastructure are provided in \cref{ExperimentSetUp}. For baselines, the results are taken from \cite{JacobiConv}. 

\subsection{Performance comparison}

The average results of running $10$ times on the node classification task are reported in \cref{Table_results}, where accuracy is used as the evaluation metric with a $95\%$ confidence interval. Both CoDeSGC-CP and -Tucker outperform all baselines on 8 out of 10 datasets. Especially, CoDeSGC-Tucker achieves significant performance gains over the SOTA method JacobiConv on several graphs, e.g., Actor and Squirrel, respectively with $4.8\%$ and $11.2\%$ improvements, which demonstrate the superiority of our models. But on two tiny graphs Texas and Cornell, our methods beat only two multi-layer GNNs, i.e., ChebNet and GCN. A possible reason is that our models are prone to overfitting the tiny graphs, since our models are more complex than these models that are derived from simpler coefficient decomposition, and we have no any constrains on the decomposition factors of the coefficients and the hyperparameter optimization on validation sets by Optuna \cite{akiba2019optuna}.      

\subsection{Ablation Analysis}
As mentioned before, coefficient decomposition underlain in JacobiConv is CP decomposition with $\boldsymbol{P=I}$. We notice that two important variations of Tucker decomposition, Tucker2 and Tucker1, which can simplify the CoDeSGC-Tucker. The Tucker2 decomposition sets one of the three factor matrices to be an identity matrix, and the Tucker1 decomposition sets two of them to be the identity matrix. Obviously, both of them have three versions. Here, we simply consider Tucker2 as setting $\boldsymbol{C=I}$ and Tucker1 as setting $\boldsymbol{C=P=I}$. Additionally, we consider testing hybrid GNNs, which contain an extra linear layer followed by a nonlinear activation to convert node features into input signals, while linear GNNs directly treat node features as input signals. For simplicity, we call hybrid GNNs "H-$\phi$" and linear GNNs "L-$\phi$", where $\phi\in \{\text{Tucker2, Tucker1, Tucker, CP}\}$. All models are tuned in the same way to ensure fairness, and their implementation details are provided in \cref{ExperimentSetUp}.  

The average results of these models on the node classification task are reported in \cref{Table_ablation}. L-Tucker performs best among linear GNNs on most datasets, but H-Tucker fails to maintain the superiority of the Tucker decomposition in most cases compared with other hybrid GNNs, which show that after adding nonlinearity, consequent extra parameters may cause overfitting. This is why the  simpler coefficient decomposition (Tucker1 and Tucker2) leads to better performance among hybrid GNNs.     

\section{Conclusion}
In this paper, we investigate existing spectral GNNs' architecture and obtain a general form with polynomial coefficients that can stores in a third-order tensor. Stripping the selection of polynomial basis and graph matrix, we find most spectral GNNs contain the spetral convolution block under the general form, which can be derived by performing different decomposition operation on the coefficient tensor. With this perspective, we propose to perform tensor decomposition, CP and Tucker, on the coefficient tensor, and obtain two corresponding spectral GNN architectures. Experimental results show that our proposed models perform better than state-of-the-art method JacobiConv on 8 out of 10 real-world datasets.   

\bibliographystyle{unsrt}  
\bibliography{references}  

\newpage
\appendix

\section{Jacobi Basis and Computing Message Propagation.}
\label{JacobiBasis}
The Jacobi basis has the following form \cite{JacobiConv}.
\begin{equation}
\label{Eq_19}
    \begin{aligned}
        P_0(s)&=1,\\
        P_1(s)&=\frac{a-b}{2}+\frac{a+b+2}{2}s.
    \end{aligned}
\end{equation}
For $k\geq 2$,
\begin{equation}
\label{Eq_20}
    P_k(s)=(\theta_ks+\theta^\prime_k)P_{k-1}(s)-\theta^{\prime\prime}_kP_{k-2}(s),
\end{equation}
where
\begin{equation}
\label{Eq_21}
    \begin{aligned}
        \theta_k &= \frac{(2k+a+b)(2k+a+b-1)}{2k(k+a+b)},\\
        \theta^\prime_k&=\frac{(2k+a+b-1)(a^2-b^2)}{2k(k+a+b)(2k+a+b-2)},\\
        \theta^{\prime\prime}_k&=\frac{(k+a-1)(k+b-1)(2k+a+b)}{k(k+a+b)(2k+a+b-2)}.
    \end{aligned}
\end{equation}
With the recursion formula of Jacobi basis shown above, we can compute $\{\boldsymbol{V}_i\}_{i=0}^{k}$ in \cref{alg_CP} and $\{\boldsymbol{\mathcal{V}}_i\}_{i=0}^{k}$ in \cref{alg_Tucker}. We here take an example using $\{\boldsymbol{V}_i\}_{i=0}^{k}$.
\begin{equation}
\label{Eq_22}
    \begin{aligned}
        \boldsymbol{V}_0&=P_0(\boldsymbol{S})\boldsymbol{H}=\boldsymbol{H},\\
        \boldsymbol{V}_1&=P_1(\boldsymbol{S})\boldsymbol{H}=\frac{a-b}{2}\boldsymbol{H}+\frac{a+b+2}{2}\boldsymbol{SH}=\frac{a-b}{2}\boldsymbol{V}_0+\frac{a+b+2}{2}\boldsymbol{SV}_0.
    \end{aligned}
\end{equation}
For $k\geq 2$,
\begin{equation}
\label{Eq_23}
    \begin{aligned}
        \boldsymbol{V}_k=P_k(\boldsymbol{S})\boldsymbol{H}&=\theta_k\boldsymbol{S}P_{k-1}(\boldsymbol{S})\boldsymbol{H}+\theta^\prime_kP_{k-1}(\boldsymbol{S})\boldsymbol{H}-\theta^{\prime\prime}_kP_{k-2}(\boldsymbol{S})\boldsymbol{H}=\theta_k\boldsymbol{S}\boldsymbol{V}_{k-1}+\theta^\prime_k\boldsymbol{V}_{k-1}-\theta^{\prime\prime}_k\boldsymbol{V}_{k-2}
    \end{aligned}
\end{equation}
One can see that the most burden in each step to compute $\boldsymbol{V}_k$ for $k=1,2,\cdots,K$ is the product of the sparse matrix $\boldsymbol{S}$ and $\boldsymbol{V}_{k-1}$. Accordingly, it is reasonable that we represent each step of the iterative propagation process as $\boldsymbol{V}_k=\Psi(\{\boldsymbol{V}_i\}_{i=0}^{k-1}),\boldsymbol{S})$. 

\section{Overview of Additional Hybrid GNNs}
\label{OtherHybrid}
\paragraph{APPNP.} APPNP \cite{APPNP} is an propagation scheme derived from PageRank, which can be formulated as:
\begin{equation}
\label{Eq_24}
    \begin{aligned}
        \boldsymbol{Z}_0&=\boldsymbol{H}=f_{\theta}(\boldsymbol{F}),\\
        \boldsymbol{Z}_{k+1}&=(1-\alpha)\boldsymbol{\tilde{A}}\boldsymbol{Z}_{k}+\alpha\boldsymbol{H},\\
        \boldsymbol{Y}&=(1-\alpha)\boldsymbol{\tilde{A}}\boldsymbol{Z}_{K-1}+\alpha\boldsymbol{H}
        \end{aligned}
\end{equation}
where $f_{\theta}()$ is a neural network usually implemented as two linear layers that are bridged by a nonlinear activation. The filtered graph signals $\boldsymbol{Y}$ will be fed into a Softmax to obtain the predictions for node labels. Recall that $\boldsymbol{F}$ is the node feature, as the aforesaid analysis for GPR-GNN \cite{GPR_GNN}, we integrate the last linear layer with the follow-up propagation process, then the model architecture can be converted to:
\begin{equation}
\label{Eq_25}
    \begin{aligned}
        &\boldsymbol{X}=f^\prime_{\theta^\prime}(\boldsymbol{F}), \\
        \boldsymbol{Y}=(1-\alpha)^{K}\boldsymbol{\tilde{A}}^K\boldsymbol{X}\boldsymbol{W}+\sum_{k=0}^{K-1}\alpha&(1-\alpha)^k\boldsymbol{\tilde{A}}^k\boldsymbol{XW}=\sum_{k=0}^{K}(1-\alpha)^k\alpha^{1-\delta_{kK}}\boldsymbol{\tilde{A}}^k\boldsymbol{XW},\\
        =\big\Vert_{j=1}^{J}&\sum_{k=0}^{K}\sum_{i=1}^{I}(1-\alpha)^k\alpha^{1-\delta_{kK}}w_{ij}\boldsymbol{\tilde{A}}^k\boldsymbol{x}_i
    \end{aligned}
\end{equation}
where $f^\prime_{\theta^\prime}()$ is the nonlinear part in $f_{\theta}$, usually containing a linear layer and a nonlinear activation if $f_{\theta}()$ follows the common implementation \cite{APPNP,ChebNetII,BernNet} of tow linear layers connected with a nonlinear activation, then $\boldsymbol{W}$ is the weight of the last linear layer in $f_{\theta}()$ and we omit the bias here. One can find that if taking $\boldsymbol{X}$ as the input signals, the convolution part (the second formula in \cref{Eq_25}) is under the unified convolution layer described in \cref{Eq_6} with $\boldsymbol{S=\boldsymbol{\tilde{A}}}$, $P_k(\boldsymbol{S})=\boldsymbol{S}^k$ and $w_{ij}^k=(1-\alpha)^k\alpha^{1-\delta_{kK}}w_{ij}$. Note that $\alpha$ is pre-defined, APPNP indeed share the same linear relationship among the orders of each filter and can learn any filters, since $w_{ij}$ are learnable.

\paragraph{BernNet and ChebNetII.} BernNet \cite{BernNet} aims at learning coefficients in a rational range and ChebNetII extends ChebNet \cite{ChebNet} by the inspiration from Chebyshev interpolation. Their implementation follows APPNP and GPR-GNN that decouple feature propagation and transformation:
\begin{equation}
    \label{Eq_26}
    \boldsymbol{Y}=\sum_{k=0}^K\alpha_kP_k(\boldsymbol{S})f_{\theta}(\boldsymbol{F})
\end{equation}
In the same way, we decouple the nonlinear part $f^\prime_{\theta^\prime}()$ from $f_{\theta}()$ and take $\boldsymbol{X}=f^\prime_{\theta^\prime}(\boldsymbol{F})$ as the input signals for the convolution part, which can be reformulated under the form of \cref{Eq_6}:
\begin{equation}
    \label{Eq_27}
    \boldsymbol{Y}=\big\Vert_{j=1}^{J}\sum_{k=0}^{K}\sum_{i=1}^{I}\alpha_kw_{ij}P_k(\boldsymbol{S})\boldsymbol{x}_i
\end{equation}
where $w_{ij}$ is the weight of the last linear layer in $f_{\theta}$ and we omit the its bias here. It is easy to see that both of them perform a decomposition $w_{ij}^k=\alpha_kw_{ij}$. The differences are that ChebNetII sets $\alpha_k=\frac{2}{K+1}\sum_{l=0}^K\gamma_lT_k(x_l)$ and $P_k(\boldsymbol{S})=T_k(\boldsymbol{S})$ with $\boldsymbol{S=L-I}$ (setting $\lambda^\ast=2$ in ChebNet \cref{Eq_3}) where $\gamma_l$ are learnable parameters and $x_l=\cos\left(\frac{l+0.5}{K+1}\pi\right)$ are called Chebyshev nodes of $T_{K+1}$; and BernNet adopts Bernstein basis $P_k(\boldsymbol{S})=\tbinom{K}{k}(1-\boldsymbol{S})^{K-k}\boldsymbol{S}^k$ with $\boldsymbol{S}=\frac{\boldsymbol{L}}{2}$ and keeps $\alpha_k$ learnable.

\paragraph{FavardGNN.} FavardGNN \cite{Optbasis} provides an approach to achieving a learnable orthogonal basis based on Favard's Theorem. The Favard basis has the following form:
\begin{equation}
\label{Eq_28}
    \begin{aligned}
        P_0(s)&=1/\sqrt{\beta_0},\\
        P_1(s)&=(s-\gamma_0)P_0(s)/\sqrt{\beta_1}.
    \end{aligned}
\end{equation}
For $k\geq 2$,
\begin{equation}
\label{Eq_29}
    P_k(s)=\left((s-\gamma_{k-1})P_{k-1}(s)-\sqrt{\beta_{k-1}}P_{k-2}(s)\right)/\sqrt{\beta_{k}},
\end{equation}
where $\sqrt{\beta_k}>0$ and $\gamma_k$ are learnable parameters. FavardGNN first converts the node features $\boldsymbol{F}$ into the graph signals $\boldsymbol{X}$ by a linear layer followed by a nonlinear activation. Next, it follows JacobiConv to filter the signals with an individual filter for each input channel and then the filtered signals are fed into a linear prediction head followed by a Softmax to get the label predictions. We can describe the FavardGNN's architecture as follows:
\begin{equation}
    \label{Eq_30}
    \begin{aligned}
        \boldsymbol{X}&=\sigma(\boldsymbol{FW}^{\prime}), \\
        \boldsymbol{Z}=\big\Vert_{i=1}^{I}&\sum_{k=0}^{K}\alpha_{ki}P_k(\boldsymbol{S})\boldsymbol{x}_i, \\
        \boldsymbol{Y} &= \boldsymbol{ZW}
    \end{aligned}
\end{equation}
where $\boldsymbol{W}^\prime$ is the weight of signal transformation layer with nonlinear activation $\sigma$, $\boldsymbol{W}$ is the weight of the prediction head and $\boldsymbol{Y}$ is the raw predictions before Softmax. Here we omit the bias used in the two linear layers for simplicity. Note that $\boldsymbol{Z}$ directly enters into the prediction head without modification from nonlinear activation. We as before take the feature propagation (the second formula in \cref{Eq_30}) and the prediction head as an whole:

\begin{equation}
    \label{Eq_31}
        \boldsymbol{Y} = \left(\big\Vert_{i=1}^{I}\sum_{k=0}^{K}\alpha_{ki}P_k(\boldsymbol{S})\boldsymbol{x}_i\right)\boldsymbol{W}=\big\Vert_{j=1}^{J}\sum_{k=0}^{K}\sum_{i=1}^{I}\alpha_{ki}w_{ij}P_k(\boldsymbol{S})\boldsymbol{x}_i
\end{equation}
Evidently, \cref{Eq_31} describes a convolution layer under the form of \cref{Eq_6} with a decomposition $w_{ij}^k=\alpha_{ki}w_{ij}$. As for the selection for propagation matrix $\boldsymbol{S}$ and the polynomial basis $P_k()$, FavardGNN follows JacobiConv to set $\boldsymbol{S=\boldsymbol{I-L}}$ and adopts the Favard basis described in \cref{Eq_28} and \cref{Eq_29}. It considers an individual filter for each input channel, and therefore, we can write the propagation process in matrix form as follows:
\begin{equation}
\label{Eq_32}
    \begin{aligned}
        \boldsymbol{V}_0=P_0(\boldsymbol{S})\boldsymbol{X}&=\boldsymbol{X}\operatorname{diag}(\boldsymbol{q}_0)^{-1},\\
        \boldsymbol{V}_1=P_1(\boldsymbol{S})\boldsymbol{X}&=\left(\boldsymbol{SV}_0-\boldsymbol{V}_0\operatorname{diag}(\boldsymbol{r}_0)\right)\operatorname{diag}(\boldsymbol{q}_1)^{-1}.
    \end{aligned}
\end{equation}
For $k\geq 2$,
\begin{equation}
\label{Eq_33}
\boldsymbol{V}_k=P_k(\boldsymbol{S})\boldsymbol{X}=\left(\boldsymbol{SV}_{k-1}-\boldsymbol{V}_{k-1}\operatorname{diag}(\boldsymbol{r}_{k-1})-\boldsymbol{V}_{k-2}\operatorname{diag}(\boldsymbol{q}_{k-1})\right)\operatorname{diag}(\boldsymbol{q}_k)^{-1},
\end{equation}
where $\boldsymbol{q}_k$ and $\boldsymbol{r}_k$ store multi-channel $\beta_k$ and $\gamma_k$. 
It is also easy to see that FavardGNN's architecture would exactly be the same as CoDeSGC-CP, 
if removing the nonlinear activation in the first layer. But, also as FavardGNN uses nonlinear activation, its first layer plays the same role as $f^\prime_{\theta^\prime}()$ in APPNP, GPR-GNN, BernNet and ChebNetII, and hence it is a hybrid GNN in our taxonomy.

\section{Experimental Settings}
\label{ExperimentSetUp}
\paragraph{The Statistics of Datasets.} The statistics of the $10$ datsets used in this work are summarized in \cref{Table_datasets}.

\begin{table*}[h]
\caption{Statistics of the used datasets.}
\label{Table_datasets}
\vskip 0.15in
\begin{center}
\begin{tabular}{lccccccccccr}
\toprule
 & Cora & CiteSeer& PubMed & Computers & Photo & Chameleon & Squirrel & Actor &Texas & Cornell \\
\midrule 
Nodes & 2708 & 3327& 19717 & 13752 & 7650 & 2277 & 5201 & 7600 &183 & 183 \\
Edges & 5278 & 4552& 44324 & 245861 & 119081 & 31371 & 198353 & 26659 &279 & 277  \\
Features & 1433 & 3703& 500 & 767 & 745 & 2325 & 2089 & 932 &1703 & 1703 \\
Classes & 7 & 6& 5 & 10 & 8 & 5 & 5 & 5 &5 & 5 \\

\bottomrule
\end{tabular}
\end{center}
\vskip -0.1in
\end{table*}

\paragraph{Computing Infrastructure.} We leverage Pytorch Geometric and Pytorch for model implementation. All experiments are conducted on an NVIDIA RTX 3090 GPU with memory of 24G on a Linux server.

\paragraph{Model hyperparameter} 
In all linear models, we directly take node features as input signals $\boldsymbol{X}$. For all hybrid models, we use one linear layer followed by a ReLU activation to transform node features into input signals. For all models, input signals are converted into output signals $\boldsymbol{Y}$ through CoDeSGC-CP (\cref{alg_CP}) or CoDeSGC-Tucker (\cref{alg_Tucker}), which are then fed into Softmax to obtain predictions. JacobiConv uses different learning rates and weight decay for linear transformation weights and propagation coefficients. Thereby, we also assign different learning rates and weight decay to all decomposition factors $\{\boldsymbol{C},\boldsymbol{b}_C\}$, $\{\boldsymbol{P}, \boldsymbol{b}_P\}$, $\boldsymbol{M}$ and $\{\boldsymbol{G}_{(1)},\boldsymbol{b}_{G}\}$. For all hybrid models, the bottom feature transformation layer also has individual learning rate and weight decay. Different dropout rates are used for node features (hybrid models), input signals (hybrid and linear models), linear transformations $\boldsymbol{C}$ (CoDeSGC-CP and -Tucker) and $\boldsymbol{G}_{(1)}$ (CoDeSGC-Tucker), and propagation step $\boldsymbol{Z}$. We fix $R=32$ in CoDeSGC-CP and $P=Q=32$ in CoDeSGC-Tucker. We treat $R$ in CoDeSGC-Tucker as a hyperparameter and tune it in a range from $\{4,8,16,32\}$. For all abovementioned learning rates, weight decay and dropout rates, as well as the hyperparameters $a,b$ in Jacobi basis, we follow JacobiConv to set the search ranges. Concretely, we select learning rates from $\{0.0005,0.001,0.005,0.01,0.05\}$, weight decay from $\{0.0,5e-5,1e-4,5e-4,1e-3\}$, dropout rates from $\{n/10\}_{n=0}^9$,$a$ from $\{-1.0+0.25n\}_{n=0}^{12}$ and $b$ from $\{-0.5+0.25n\}_{n=0}^{10}$. For hybrid models, we set the dimensions of hidden states after the first layer as $64$, also hybrid models' channel number of input signals. The Adam optimizer is used for training all models. We initialize all parameters through default random initialization, i.e., following a uniform distribution in the interval $[-\frac{1}{\sqrt{F_{in}}},\frac{1}{\sqrt{F_{in}}}]$. Here, $F_{in}=K+1$ for $\boldsymbol{M}$ and for other parameters, $F_{in}$ is the input dimensions. Note that, we do not include any tricks in model training stage.

\end{document}